\pdfoutput=1

\documentclass[11pt]{article}

\usepackage[]{EMNLP2022}

\usepackage{times}
\usepackage{latexsym}

\usepackage[T1]{fontenc}

\usepackage[utf8]{inputenc}

\usepackage{microtype}

\usepackage{booktabs} 

\definecolor{mark}{RGB}{230, 242, 255}
\definecolor{mark2}{RGB}{230, 230, 255}

\newcommand{\chick}{\raisebox{-2pt}{\includegraphics[width=0.15in]{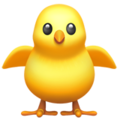}}\textsc{model}}
\newcommand{\rooster}{\raisebox{-2pt}{\includegraphics[width=0.15in]{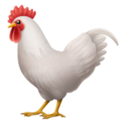}}\textsc{model}}

\usepackage{inconsolata}
\usepackage{graphicx} 
\graphicspath{ {figures/} }
\usepackage{xcolor}
\usepackage{pifont}

\usepackage{makecell} 
\usepackage{multirow} 

\title{Lexical Generalization Improves with Larger Models and Longer Training}
\author{Elron Bandel$^{1,2}$ \qquad Yoav Goldberg$^{1,3}$ \qquad Yanai Elazar$^{3,4}$   \\
  $^1$Computer Science Department, Bar Ilan University\\
  $^2$IBM Research \qquad
  $^3$Allen Institute for Artificial Intelligence \\
  $^4$Paul G. Allen School of Computer Science \& \\ Engineering, University of Washington \\
  \texttt{elron.bandel@gmail.com}}
\begin{document}
\maketitle
\begin{abstract}


While fine-tuned language models perform well on many tasks, they were also shown to rely on superficial surface features such as lexical overlap. Excessive utilization of such heuristics can lead to failure on challenging inputs. We analyze the use of lexical overlap heuristics in natural language inference, paraphrase detection, and reading comprehension (using a novel contrastive dataset),
and find that larger models are much less susceptible to adopting lexical overlap heuristics. We also find that longer training leads models to abandon lexical overlap heuristics. Finally, we provide evidence that the disparity between models size has its source in the pre-trained model.\footnote{Code and data are available at: \url{https://github.com/elronbandel/lexical-generalization}.}
\end{abstract}

\section{Introduction}

Pretrained Language Models (PLMs) dramatically improved the performances on a wide range of NLP tasks, resulting in benchmarks claiming to track progress of language understanding, like GLUE \cite{wang-etal-2018-glue} and SQuAD \cite{rajpurkar-etal-2016-squad} to be ``solved''.
However, many works show these models to be brittle and generalize poorly to ``out-of-distribution examples'' \cite{naik-etal-2018-stress,mccoy-etal-2019-right,gardner-etal-2020-evaluating}.

\begin{figure}[!t]
    \centering
    \includegraphics[scale=0.39]{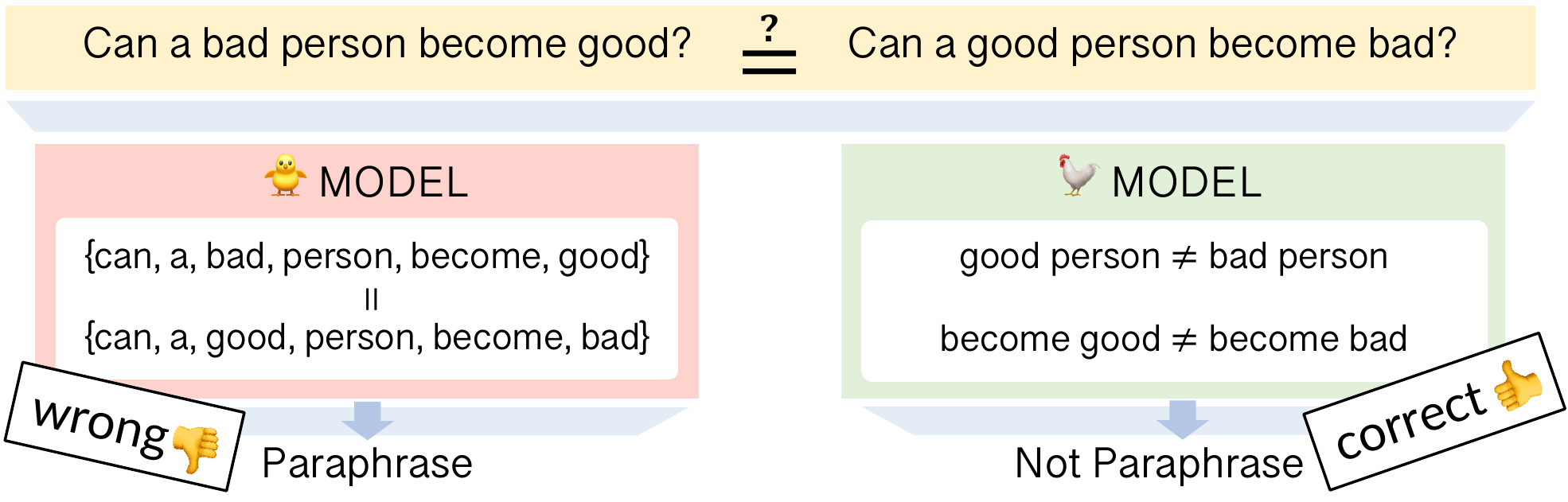}
    \caption{Illustration of the difference in the behavior of a too small, or briefly trained paraphrase detection \chick{}, that relies on lexical overlap heuristic to make a (wrong) prediction. In contrast, \rooster{} is larger or is trained for longer, and does not rely on lexical overlap and predicts correctly.}
    \label{fig:exm}
\end{figure}

One of the reasons for the poor generalization is models adopting superficial heuristics from the training data, such as, \textit{lexical overlap}: simple match of words between two textual instances. For instance, a paraphrase-detection model that makes use of these heuristics may determine that two sentences are paraphrases of each other by simply comparing the bags-of-words of these sentences. While this heuristic sometimes works,
it is also often wrong, as demonstrated in Figure \ref{fig:exm}. 
Indeed, while such heuristics are effective for solving in-domain large datasets, different 
tests expose that models often rely on these heuristics, and thus they are \textit{right for the wrong reasons} \cite{mccoy-etal-2019-right,zhang-etal-2019-paws}.

Since, different works tackled these problems, and propose different algorithms and specialized methods for reducing the use of such heuristics, and improve models’ generalization \cite{he-etal-2019-unlearn,utama-etal-2020-towards,Moosavi2020ImprovingRB, tu-etal-2020-empirical, liu-etal-2022-wanli}.

In this work, we link the adoption of lexical overlap heuristic to size of PLMs and to the number of iterations in the fine-tuning process. We show that \textbf{a lot of the benefit from the above methods can be achieved simply by using larger PLMs and finetuning them for longer}.

We show that larger PLMs, and longer trained models are much less prone to rely on lexical overlap, \textbf{despite not being manifested on standard validation sets}.
We validate these findings on three widely used PLMs: BERT \cite{devlin-etal-2019-bert}, RoBERTa \cite{Liu2019RoBERTaAR}, and ALBERT \cite{lan2019albert} and three tasks: Natural Language Inference (NLI), Paraphrase Detection (PD), and Reading Comprehension (RC). For RC, we collect a new contrastive test-set - \textsc{ALSQA}, based on SQuAD2.0 \cite{rajpurkar-etal-2018-know}, while controlling on the lexical overlap between the texts and the questions,
containing 365 examples.
\section{Related Work}
This work relates to a line of research using behavioral methods for understanding model behavior \cite{ribeiro-etal-2020-beyond,Elazar2021AmnesicPB,vig2020causal}, and more specifically regarding the extent in which specific heuristics being used by models for prediction \cite{poliak2018hypothesis,naik-etal-2018-stress,mccoy-etal-2019-right,contrastive-explanations,winograd-elazar}.
%
%
%
%
The use of heuristics such as lexical overlap also typically point on the lexical and under sensitivity of models \cite{welbl-etal-2020-undersensitivity}, studied in different setups \cite{iyyer-etal-2018-adversarial,jia-liang-2017-adversarial,gan-ng-2019-improving,misra-etal-2020-exploring,Belinkov2018SyntheticAN,ribeiro-etal-2020-beyond,ebrahimi-etal-2018-hotflip, bandel-etal-2022-quality}.

Our work suggests that the size of PLMs affects their \textit{inductive bias} \cite{HAUSSLER1988177} towards the preferred strategy.
Studies of inductive biases in NLP has gained attention recently \cite{dyer-etal-2016-recurrent,Battaglia2018RelationalIB,dhingra-etal-2018-neural,ravfogel-etal-2019-studying,McCoy2020DoesSN}. While \citet{warstadt-etal-2020-learning} studied the effect of the amount of pre-training data on linguistic generalization, we show that additional training iterations and larger models affects their inductive biases with respect to the use of lexical overlap heuristics.

Finally, \citet{tu-etal-2020-empirical} show that the adoption of heuristics can be explained by the ability to generalize from a few examples that aren't aligned with the heuristics. They show that larger model size and longer fine-tuning can marginally increase the ability of the model to generalize from minority groups, an insight also discussed also by \citet{djolonga}. Our focus of lexical heuristics exclusively, together with robust fine-grained experimental setup concludes, unlike \citet{tu-etal-2020-empirical}, that the change in lexical heuristic adoption behavior is consistent and non marginal.

\section{Measuring Reliance on Lexical Overlap}
\label{sec:lexical-overlap}

\begin{figure*}[t!]
\centering
\resizebox{1\textwidth}{!}{%
\begin{tabular}{llll}
\toprule
\textbf{Dataset} &  \textbf{Text1} &\textbf{Text2} & \textbf{Label}\\
\midrule
HANS & The banker near the judge saw the actor. &  The banker saw the actor.  & E \, \, \ding{51} \\

  & The doctors visited the lawyer. & The lawyer visited the doctors.  & NE \ding{55}\\
\midrule
PAWS  & What should I prefer study or job? & What should I prefer job or study? & P \, \, \ding{51}\\

  & Can a bad person become good ? & Can a good person become bad?  & NP \ding{55} \\
\midrule

ALSQA & \multirow{3}{*}{\makecell[l]{..."downsize" revision of \colorbox{mark2}{vehicle} \colorbox{mark2}{categories}.By \colorbox{mark2}{1977}, GM\'s \\ \colorbox{mark}{full}-\colorbox{mark}{sized} \colorbox{mark}{cars} reflected the crisis. By 1979, virtually all  \\"\colorbox{mark}{full} -\colorbox{mark}{size}" \colorbox{mark}{American} \colorbox{mark}{cars} had \colorbox{mark}{shrunk}, featuring \colorbox{mark}{smaller} \\ engines and \colorbox{mark}{smaller} outside dimensions. \colorbox{mark2}{Chrysler} ended \\ production of their \colorbox{mark}{full}-\colorbox{mark}{sized} luxury  sedans at the end  of the \\ 1981 model \colorbox{mark}{year}...}} & \multirow{3}{*}{\makecell[l]{By which\colorbox{mark}{year} did \colorbox{mark}{full} \colorbox{mark}{sized} \colorbox{mark}{American}\\ \colorbox{mark}{cars} \colorbox{mark}{shrink} to be \colorbox{mark}{smaller}? \\ \\ What \colorbox{mark2}{vehicle} \colorbox{mark2}{category} did \colorbox{mark2}{Chrysler} \\ change to in \colorbox{mark2}{1977}?}} &\multirow{3}{*}{\makecell[l]{A \enspace \ding{51}\\ \\ \\ NA \ding{55} \\ }}  \\
& & &  \\
& & &  \\
& & &  \\
& & &  \\
& & &  \\
& & &  \\[-0.7em]
\bottomrule
\end{tabular}
}
\caption{Examples from the datasets. All examples have in-pair high lexical overlap. In ALSQA examples overlapping content words colored in blue . Text1 and Text2 correspond to the premise and hypothesis in HANS, the two sentences in PAWS and the context and question in ALSQA. 
The labels for HANS pairs are either Entailment (E) or Non-Entailment (NE). For PAWS the labels are Paraphrase (P) or Non-Paraphrase (NP). For ALSQA the labels are (A) Answerable (NA) Non-Answerable. If the lexical overlap point on the label it is marked as \ding{51} means it is \emph{consistent-with-heuristic}.}

\label{tab:examples}
\end{figure*}

\paragraph{Definition: Reliance on Lexical Overlap}
We say that a model makes use of \textit{lexical heuristics} if it relies on superficial lexical cues in the text by making use of their identity rather than the semantics of the text. 

\subsection{Experimental Design}
We focus on tasks that involve some inference over text pairs.
We estimate the usage of the lexical-overlap heuristic by training a model on a \textit{standard training set} for the task, and then inspecting models' predictions on a \emph{diagnostic set}, containing high lexical overlap instances,\footnote{We define the \emph{lexical overlap} between pair of texts as the number of unique words that appear in both texts divided by the number of unique words in the shorter text. The definition for ALSQA is similar, but we consider lemmas instead of words, and ignore function words and proper names.} where half of the instances are consistent with the heuristic and half are inconsistent (e.g. the first two rows of each dataset in Figure \ref{tab:examples}, respectively).
Models that rely on the heuristic will perform well on the consistent group, and significantly worse on the inconsistent ones.

\paragraph{Metric} We define the HEURistic score (HEUR) of a model on a diagnostic set as the difference in model performance on the consistent examples, and its performance on the inconsistent examples. Higher HEUR values indicate high use of the lexical overlap heuristic (bad) while low values indicate lower reliance on the heuristic (good).

\subsection{Data}
For the \textbf{training sets} we use
the MNLI dataset \cite{williams-etal-2018-broad} for Natural Language Inference (NLI; \citealp{dagan2005pascal,bowman2015large}), the Quora Question Pairs (QQP; \citealp{Sharma2019NaturalLU}) for paraphrasing, and SQuAD 2.0 \cite{rajpurkar-etal-2018-know} for Reading Comprehension (RC).
The corresponding high-lexical overlap diagnostic sets are described below:
\paragraph{HANS}\cite{mccoy-etal-2019-right}
is an NLI dataset, designed as a challenging test set to test for reliance on different heuristics.
Models that predict entailment when based solely on one of those heuristics - will fail in half the examples where the heuristic does not hold.
We focus on the lexical overlap heuristic part, termed HANS-Overlap.

\paragraph{PAWS}\cite{zhang-etal-2019-paws}
is a paraphrase detection dataset containing paraphrase and non-paraphrase pairs with high lexical overlap. The pairs were generated by controlled word swapping and back translation, with a manually filtering validation step. 
We use the Quora Question Pairs (PAWS-QQP) part of the dataset.
\paragraph{ALSQA} To test the lexical overlap heuristic utilization in Reading Comprehension models, we create a new test set: 
Analyzing Lexically Similar QA (ALSQA).
We augment the SQuAD 2.0 dataset \cite{rajpurkar-etal-2018-know} by asking crowdworkers to generate questions with high context-overlap from questions with low overlap (These questions are paraphrases of the original questions). In the case of un-answerable questions, annotators were asked to re-write the question without changing its meaning and maintain the unanswerability reason.\footnote{Full details are provided in Appendix \ref{sec:alsqa}.}
ALSQA contains 365 questions pairs, 190 with answer and 174 without answer. Examples from the dataset are presented in Figure \ref{tab:examples}, and Appendix \ref{sec:alsqa_exm}.

\begin{figure}[t!]
    \centering
    \includegraphics[width=1.\columnwidth]{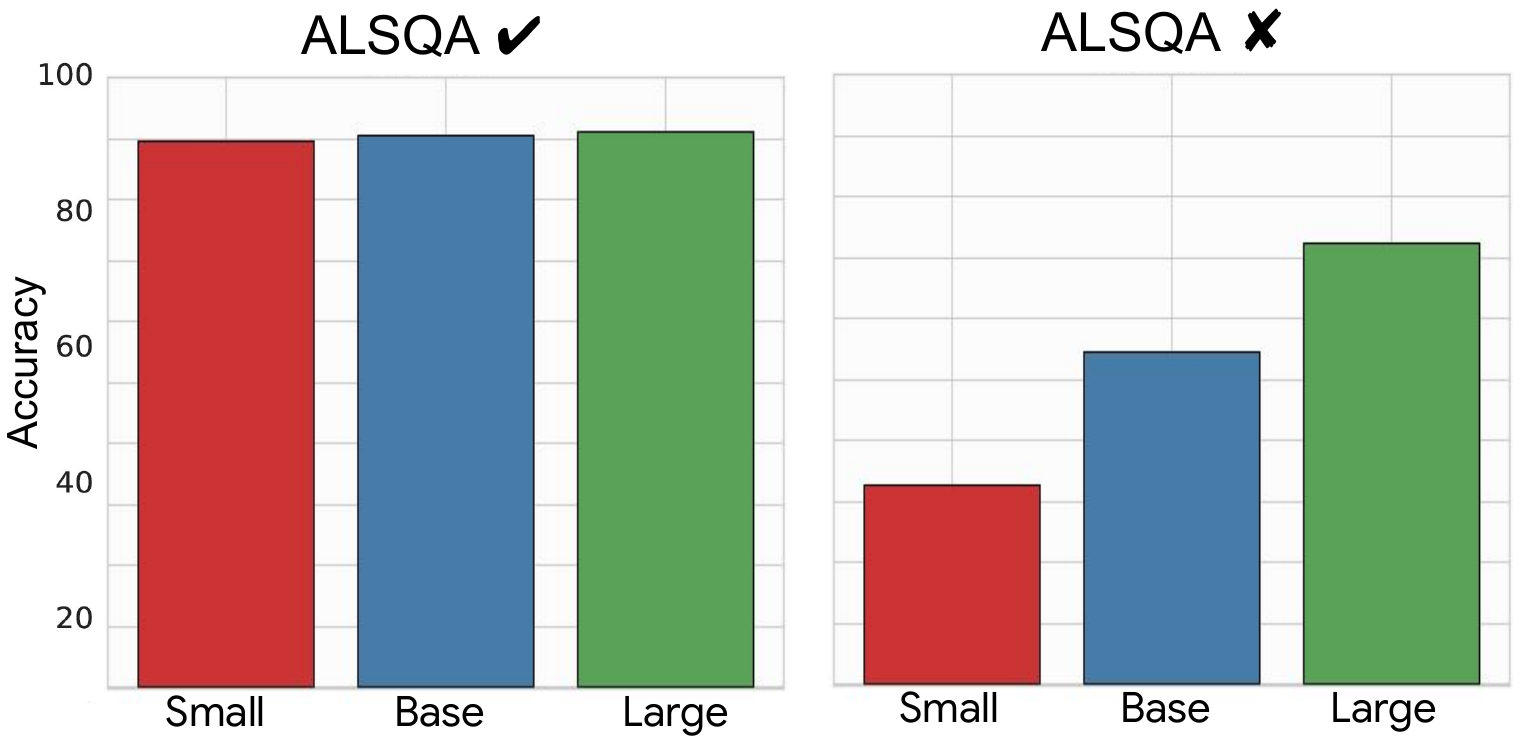}
    \caption{\textbf{"Larger is better"} Electra of different sizes perform equally on the subset of \textsc{ALSQA} that is consistent with the lexical overlap heuristic (\textbf{\ding{51}}), in the inconsistent subset (\textbf{\ding{55}}) larger models are less likely to adopt the heuristic, therefore, generalize better. }
    \label{fig:alsqa_electra}
\end{figure}

\section{Experiments and Results}
\label{sec:results}

\paragraph{Setup}

We experiment with 3 strong PLMs: BERT \cite{devlin-etal-2019-bert}, RoBERTa \cite{Liu2019RoBERTaAR} and ELECTRA \cite{Clark2020ELECTRA:}, each with a few size variants. Since performance after finetuning can vary, both for in-domain \cite{Dodge2020FineTuningPL}, and out-of-domain \cite{mccoy-etal-2020-berts}, we follow \citet{Clark2020ELECTRA:} and finetune every model six times, with different seeds and learning rates (specified in Appendix \ref{sec:train}) and report the median of these results. 
All models achieve comparable results to those reported in the literature. 
For each model we consider different stages in the training process. The \emph{early} variant is after one epoch, while the \emph{late} variant is after six epochs. 

\begin{table}[t!]
\centering
\resizebox{1\columnwidth}{!}{%
\begin{tabular}{llrrrrr}
\toprule
\textbf{Model-Stage} &\textbf{Size} & \textbf{Dev}$\uparrow$ & \textbf{\ding{51}}$\uparrow$ & \textbf{\ding{55}}$\uparrow$ & \textbf{HEUR}$\downarrow$ & \textbf{$\Delta$}\\
\midrule
\multicolumn{6}{l}{\textbf{NLI}} \\
\midrule
 BERT early & base     & 83.8 & 97.3 & 3.6 &  93.7& \\
&large     & 84.9 & 91.1 & 27.6 & 63.5& \\
\midrule
BERT late & base & 84.5 & 62.0 & 43.5 & 18.5 & -75.2\\
    &  large& 85.8 & 82.6 & 68.0 & 14.6 & -48.9\\
\midrule
RoBERTa early & base & 85.9 & 99.4	 & 9.7 &  89.7& \\
     & large & 87.8 & 99.8 &	81.4 &  18.4& \\
\midrule
RoBERTa late & base & 87.1 & 98.0 & 76.8 & 21.2 &  -68.5\\
    &  large & 89.3  & 99.6	& 92.4 & 4.2 & -14.2\\
\midrule
\multicolumn{6}{l}{\textbf{Paraphrase Detection}} \\
\midrule
RoBERTa early& base & 89.4 & 93.7	 & 9.3 & 84.4& \\
    & large & 89.1 & 95.0 &	19.0 & 76.0 & \\
\midrule
RoBERTa late & base &91.6 & 90.1 & 21.9 & 68.2 & -16.2\\
        & large & 91.9 & 94.8	& 23.9 & 70.9 & -5.1\\
\midrule
ELECTRA early & base & 90.0 & 89.3 &	17.5 & 71.8 & \\
         & large  & 90.8 & 98.4 &	23.9 & 74.5 & \\
\midrule
ELECTRA late &  base & 91.8 & 89.5	& 35.1 & 54.4 & -17.4\\
        & large  & 92.5 & 93.7 &	42.0 & 51.7 &  -22.8\\
\midrule
\multicolumn{6}{l}{\textbf{Question Answerability}}\\
\midrule
ELECTRA early & base & 83.5 & 90.8 & 53.7 & 37.1 & \\
         & large  & 90.3 & 89.5 &	71.4 & 18.1  & \\
\midrule
ELECTRA late & base & 83.5 & 90.5	& 54.6 & 35.9 & -1.2\\
        & large  & 91.0 & 91.1 & 72.3 & 18.8 &  0.7\\
\bottomrule

\end{tabular}
}
\caption{
Results on all tasks considered in this study. We include both the dev-set results on the in-domain dataset (Dev), and the HEUR columns are reported on the high lexical overlap diagnostic sets.
\ding{51} and \ding{55} refer to the subsets from the diagnostic sets that are consistent, and inconsistent with the heuristic, respectively. $\Delta$ refers to the difference in HEUR between the same model family of the same size, between the late and early stage of training (e.g. the difference between BERT-base in the late and early stage of training).
}

\label{tab:results}
\end{table}
The NLI and PD are text-pair classification tasks. 
However, RC is a span prediction task, as such, we explore two versions of this task: (1) the regular span prediction task, and (2) a text-pair classification task, where the goal is to predict if the questions is answerable or not from the text, which we call \textit{Answerability}.

\paragraph{Finding I: Larger is Better {\small(but it is not reflected on the dev set)  }}
We report the results in Table \ref{tab:results}.
Larger models perform consistently better on the lexical challenge across tasks and models:
For NLI, the \emph{early} BERT models gradually improve their HEUR score from 93.7 in the base version to 63.5 in the large version.
Similarly for PD, the early RoBERTa improves the HEUR score from 84.4 to 76.0 from the base to large versions.
The difference in the HEUR scores are remarkable, as the improvement on the standard validation set
are relatively small with respect to the improvement on HEUR. For instance, the RoBERTa-early on HANS improve its dev performance from the base to large version by 2.2\%, while HEUR improves relatively by 79.4\%.
Apart from the size trend, the absolute numbers between the tasks also differ significantly. RoBERTa-large practically ``solved'' this part of HANS, with a median accuracy of 96\%.
On the other hand, PAWS remains a much harder task, with the best performing model is ELECTRA-large that obtains 56.6\% accuracy. 

\begin{figure}[t!]
    \centering
    \includegraphics[width=1.\columnwidth]{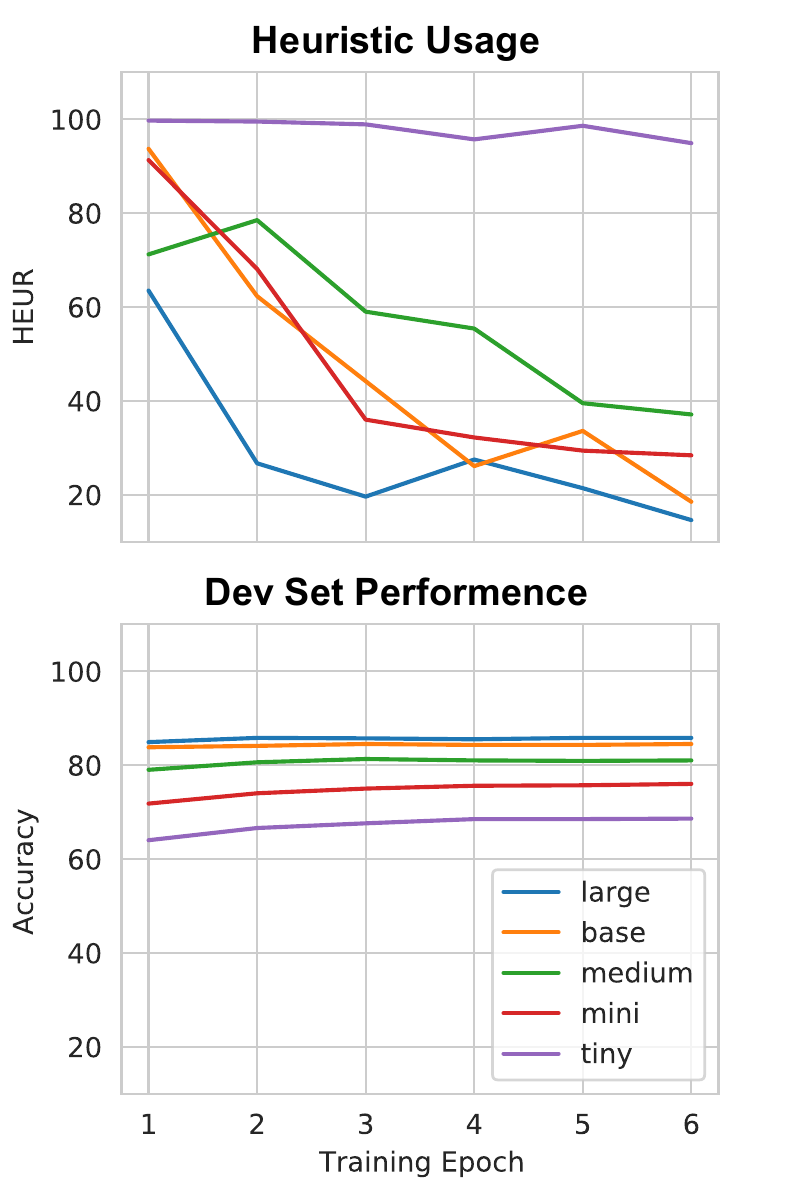}
    \caption{ \textbf{"Longer is better"} Lexical Overlap HEUR of different sizes of BERT on HANS during training compared to dev set accuracy. While dev set results remain stagnant over training, the behavior of the model changes dramatically with regards to the reliance on the lexical heuristic, as reflected in the HEUR score.}
    \label{fig:hans_bert_std} 

\end{figure}

\paragraph{Finding II: Longer is Better {\small(but it is not reflected on the dev set)  }}
Next, we compare the \emph{early} (after one iterations over the training data) and \emph{late} (after six iterations) versions of models. 
To ease comparison, the $\Delta$ column indicates the difference between the HEUR score on the same model size, at the early and late stages of training.
The differences show a constant improvement on this score from early to later versions of the trained models. Accordingly, the improvements on the standard development sets are again much smaller: for instance, on NLI using BERT-large the dev-score increases by 1\% while the HEUR score  improves relatively by 77\% (see Figure \ref{fig:hans_bert_std})\footnote{The trend seems to not hold in ELECTRA-large in Table \ref{tab:results} and both RoBERTa models in Table \ref{tab:span-prediction-results}. However, we found that in an earlier training stage (middle of the first training iteration) the HEUR scores were significantly worse, while the dev scores were similar.}
This indicates that models tend to adopt the lexical overlap heuristic at early training stages, and abandon it later on.

\paragraph{Span-Prediction Objective}
We also compare models that were trained on the span-prediction task (RC) on SQuAD 2.0 to inspect if the task objective and loss affect the usage of lexical heuristic. We follow the same training setup solely for SQuAD 2.0 with the standard span-prediction task and report the results in Appendix \ref{sec:add_res_span}.
We observe similar trends: larger or sufficiently fine-tuned models are less prone to use the overlap heuristic.
\begin{table}[t!]
\centering
\resizebox{1\columnwidth}{!}{%
\begin{tabular}{llrrrr}
\toprule
            &  &  \multicolumn{2}{c}{\textbf{High Probability}} &  \multicolumn{2}{c}{\textbf{Low Probability}} \\
      \textbf{Model} & \textbf{Size} &  \textbf{$PPL$} & $\Delta$ &  \textbf{$PPL$} & $\Delta$ \\
\midrule

    BERT & base &          2.5 &          &       11.2 &       \\
    &large &          2.4 &          -4.0\% &        9.5 &        -15.1\% \\
   \midrule
  RoBERTa &base &          2.4 &                &       14.1 &              \\
 &large &          2.2 &          -8.3\% &        9.4 &        -33.3\% \\
\bottomrule
\end{tabular}
}
\caption{
The difference between the perplexity of pre-trained models sizes on probable and improbable sentences. On the probable sentences perplexity differences are marginal while on the improbable sentences differences are  worse.
 Full results provided in Appendix  \ref{sec:add_res}. }
\label{tab:perp}
\end{table}
\paragraph{On the Source of Generalization}
Is the improvement in generalization based only on the capacity of the model being fine-tuned and the amount of fine-tuning, or are the larger pre-trained models themselves already better at generalization? 
\citet{tu-etal-2020-empirical} show that larger models generalize better on challenging datasets by learning from a low probability sub-population of the training data (where the heuristics do not hold)\footnote{Less then 0.1\% of the training data \cite{mccoy-etal-2019-right}}. However, it remains unclear if the capability to learn from improbable sub-populations, pre-exists from pre-training or emerges at fine-tuning. 
We speculate that \textbf{this capability has its source in the pre-trained model}.In the following experiment, we try and demonstrate that the larger pre-trained models are indeed better at generalization. We show that larger pre-trained models perform better on improbable sub-populations of the pre-training data, while having a similar performance as smaller models on the probable population. 
To approximate the distribution of the pre-training texts, we use an LM ensemble, and sort 3000 unseen sentences by their probability assigned by the ensemble. Doing so, we observe that larger LMs achieve considerably better  perplexity on the improbable sentences, compared to the small models (15-33\% relative improvement), while having comparable perplexity on the probable sentences (4-8\% improvement). We provide more details and the full results in Appendix \ref{sec:perplexity_app}. One possible explanation is that the larger models have higher memorization capacity \cite{Tirumala2022MemorizationWO} that can be helpful for learning from large variety of improbable sub-populations \cite{Feldman2020DoesLR}. This pre-training skill can be utilized for generalizing from improbable groups, such as the examples where the heuristic does not hold, later in training\footnote{Additionally, the link between the perplexity and the skill of learning from improbable data populations, might explain why models with better perplexity (such as RoBERTa in comparison to BERT of the same size, when PPL compared properly \cite{dudy-bedrick-2020-words}) tend to have lower HEUR scores (Table \ref{sec:results}).}.

\section{Conclusions and Discussion}





We show that (1) larger pre-trained models are less prone to make use of lexical overlap heuristics in fine-tuning; and (2) longer fine-tuning may reduce the use of such heuristics. 
These findings raise questions on the training dynamics of fine-tuned LMs \cite{saphra-lopez-2019-understanding,chiang-etal-2020-pretrained}, which should be explored in future work: What kind of inductive biases does large models posses towards better generalization abilities? What makes them adopt and abandon overlap heuristics with minimal signal from the training data? We suggest that larger models are better at learning from improbable groups in the data due to their larger capacity. To support this claim we show that larger PLMs achieve better perplexity mainly in less probable sentences.
Our work is the first to suggest that sufficiently-trained large PLMs are capable at arriving at solutions that are not wrongly reliant on lexical overlap. Such models should be used as baselines when developing techniques to alleviate the reliance of lexical heuristics \cite{he-etal-2019-unlearn,utama-etal-2020-towards,Moosavi2020ImprovingRB} and assessing progress \cite{bowman-2022-dangers}.


\section{Limitations}

This work focuses solely on the effect of the lexical overlap heuristic and how models adopt and abandon it. As such, we study in depth this heuristic and its source of origin, but we cannot make broader claims on the quality of larger models and longer training, in general. More research is needed to study the effect of these models in other scenarios and heuristics to gain better understanding of their capabilities. Additionally, this work was mostly conducted two years ahead of publication, and while the size of the models and the length of training seemed to be enough back then, by today's standards they could be enlarged. Finally, like other empirical works in this field, this work makes broad claims over a large population (the possible space of all pre-trained language models) based on observation of a small sample from it. As with other works in this field, we do believe our conclusions to be useful despite the small-sample issue.   \newpage



\section*{Acknowledgments}
This project has received funding from the European Research Council (ERC) under the European Union's Horizon 2020 research and innovation programme, grant agreement No. 802774 (iEXTRACT).
Yanai Elazar is grateful to have been supported by the PBC fellowship for outstanding PhD candidates in Data Science and the Google PhD fellowship for his PhD, where he spent most of his time on this project.




\bibliography{anthology,custom}
\bibliographystyle{acl_natbib}

\appendix


\newpage
\section{Further Details on ALSQA}
\label{sec:alsqa}


\paragraph{Lexical Overlap}
We consider three distinct classes of lexical overlap: High, Medium, and Low.
We measure it as the overlap coefficient \cite{MK2016ASO} between the bag of lematized\footnote{based on spaCy \cite{spacy}.} content words of two sentences. We define content words as words that are not named entities determiners or question words. We define lexical overlap coefficient higher than 0.85 as High, lower or equal to 0.85 and higher than 0.4 as Medium, and lower or equal than 0.4 as Low.



\paragraph{Collection Process}
To build the dataset, we sampled questions from the development set of SQuAD2.0 and calculated the lexical overlap level with their corresponding passages: High, Medium or Low. 
In unanswerable questions we considered the overlap between the question and the whole passage. In answerable questions we considered the lexical overlap between the question and the sentences containing the answer, in addition to one sentence before and after.
\begin{figure*}[t!]
    \centering
    \includegraphics[width=1.8\columnwidth]{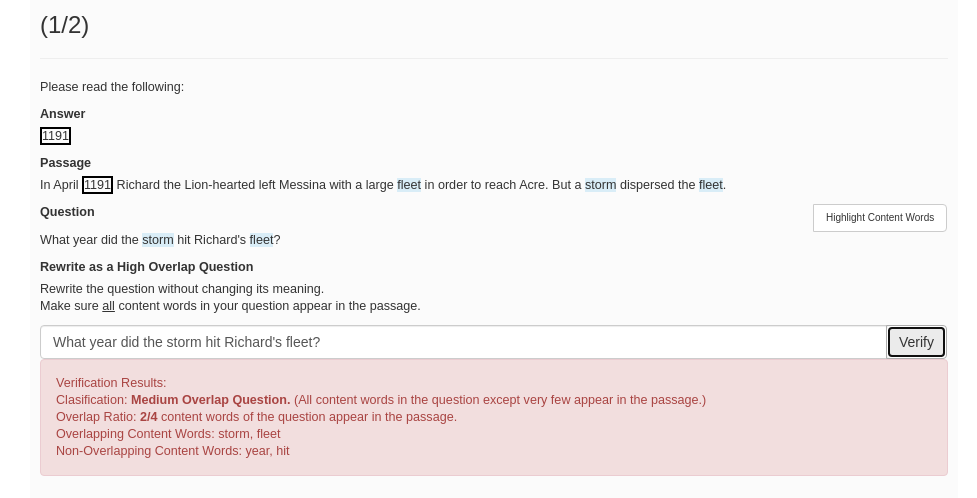}
    \caption{A screenshot of the annotation interface used for collecting ALSQA. Verification process was employed to assist the annotators. The content words are identified both in the question and the passage, and are highlighted in blue after the annotator press the `verify' button.}
    \label{fig:system}                                  
\end{figure*}
We then filtered out high and medium overlap questions. 
Given a low overlap question, answer and a passage, the annotators were asked to rewrite the question such that (1) it will maintain its meaning in the context of the passage; (2) it will fit the High overlap class. We had an interactive lexical overlap validator to verify the paraphrased question indeed have high overlap. 
We also highlighted the overlapping and non-overlapping words in the question for facilitating the annotation procedure.
In case the annotators did not manage to rewrite questions that satisfy the requirements they could provide an explanation and skip the question. Only less than 10\% of the questions were skipped by annotators, and we removed these questions from the final dataset.

\paragraph{Annotators Preparation}
Following \citet{roit-etal-2020-controlled,pyatkin-etal-2020-qadiscourse}, we trained 8 annotators on a small amount of examples and gave them detailed feedback on their annotations. Then, we tested them on another small batch and proceeded with the dataset collection process only with annotators with high success rates on the test (5 annotators in total). Finally we added the annotations of an annotator to the final dataset only if we manually approved 95\% of the annotations in 25\% of the annotator work. We paid 0.4\$ for each annotation including the ones we rejected. 

\section{Additional ALSQA Examples}
\label{sec:alsqa_exm}
In this section, we give a few more examples from the \textsc{ALSQA} dataset. Each example contains the original SQuAD2.0 context paragraph, a human generated question with high overlap with this paragraph, and the SQuAD2.0 answer for the question if it exists.

\begin{center}
\begin{tabular}{|l|}
\hline
\textit{Example 1} \\
\hline
\textbf{Context} \\
\begin{minipage}[t]{0.944\columnwidth}%
To the east is the Colorado Desert and the Colorado River at the border with Arizona, and the Mojave Desert at the border with the state of Nevada. To the south is the Mexico–United States border. \\
\end{minipage}\tabularnewline
\hline
\textbf{Question} \\
\begin{minipage}[t]{0.944\columnwidth}%
What is at the south border with Arizona?
\end{minipage}\tabularnewline
\hline
\textbf{Answer} \\
\begin{minipage}[t]{0.944\columnwidth}%
None
\end{minipage}\tabularnewline
\hline
\end{tabular}
\end{center}
\begin{center}
 \begin{tabular}{|l|}
 \hline
\textit{Example 2} \\
\hline
\textbf{Context} \\
\begin{minipage}[t]{0.944\columnwidth}%
To the east is the Colorado Desert and the Colorado River at the border with Arizona, and the Mojave Desert at the border with the state of Nevada. To the south is the Mexico–United States border.
\\
\end{minipage}\tabularnewline
\hline
\textbf{Question} \\
\begin{minipage}[t]{0.944\columnwidth}%
What borders with the state of Nevada?
\end{minipage}\tabularnewline
\hline
\textbf{Answer} \\
\begin{minipage}[t]{0.944\columnwidth}%
Mojave Desert
\end{minipage}\tabularnewline
\hline
\end{tabular}
\end{center}

\begin{center}
 \begin{tabular}{|l|}
 \hline
\textit{Example 3} \\
\hline
\textbf{Context} \\
\begin{minipage}[t]{0.944\columnwidth}%
The San Bernardino-Riverside area maintains the business districts of Downtown San Bernardino, Hospitality Business/Financial Centre, University Town which are in San Bernardino and Downtown Riverside.
\\
\end{minipage}\tabularnewline
\hline
\textbf{Question} \\
\begin{minipage}[t]{0.944\columnwidth}%
What area maintains the districts University Town and business districts downtown?\\
\end{minipage}\tabularnewline
\hline
\textbf{Answer} \\
\begin{minipage}[t]{0.944\columnwidth}%
Riverside
\end{minipage}\tabularnewline
\hline
\end{tabular}
 \begin{tabular}{|l|}
 \hline
\textit{Example 4} \\
\hline
\textbf{Context} \\
\begin{minipage}[t]{0.944\columnwidth}%
College sports are also popular in southern California. The UCLA Bruins and the USC Trojans both field teams in NCAA Division I in the Pac-12 Conference, and there is a longtime rivalry between the schools.
\\
\end{minipage}\tabularnewline
\hline
\textbf{Question} \\
\begin{minipage}[t]{0.944\columnwidth}%
The field teams from southern California were both from schools in sports in which NCAA group? \\
\end{minipage}\tabularnewline
\hline
\textbf{Answer} \\
\begin{minipage}[t]{0.944\columnwidth}%
Division I
\end{minipage}\tabularnewline
\hline
\end{tabular}
\end{center}

\begin{center}
 \begin{tabular}{|l|}
 \hline
\textit{Example 5} \\
\hline
\textbf{Context} \\
\begin{minipage}[t]{0.944\columnwidth}%
Even before the Norman Conquest of England, the Normans had come into contact with Wales. Edward the Confessor had set up the aforementioned Ralph as earl of Hereford and charged him with defending the Marches and warring with the Welsh. In these original ventures, the Normans failed to make any headway into Wales.\\
\end{minipage}\tabularnewline
\hline
\textbf{Question} \\
\begin{minipage}[t]{0.944\columnwidth}%
Who made Edward the Confessor into an Earl?\\
\end{minipage}\tabularnewline
\hline
\textbf{Answer} \\
\begin{minipage}[t]{0.944\columnwidth}%
None
\end{minipage}\tabularnewline
\hline
\end{tabular}
\end{center}

\begin{center}
 \begin{tabular}{|l|}
 \hline
\textit{Example 6} \\
\hline
\textbf{Context} \\
\begin{minipage}[t]{0.944\columnwidth}%
The further decline of Byzantine state-of-affairs paved the road to a third attack in 1185, when a large Norman army invaded Dyrrachium, owing to the betrayal of high Byzantine officials. Some time later, Dyrrachium—one of the most important naval bases of the Adriatic—fell again to Byzantine hands.\\
\end{minipage}\tabularnewline
\hline
\textbf{Question} \\
\begin{minipage}[t]{0.944\columnwidth}%
What one of the most important naval bases of the Adriatic-fell to the Normans?\\
\end{minipage}\tabularnewline
\hline
\textbf{Answer} \\
\begin{minipage}[t]{0.944\columnwidth}%
None
\end{minipage}\tabularnewline
\hline
\end{tabular}
 \end{center}

\newpage
\section{Training Details And Reproducibility}
\label{sec:train}

The finetuning was done using PyTorch finetuning code can be found on huggingface's transforemers \cite{2020HuggingFace-datasets} github repository.
For QQP and MNLI we used the following code: \url{https://github.com/huggingface/transformers/blob/main/examples/pytorch/text-classification/run\_glue.py} and for SQuAD2.0: \url{https://github.com/huggingface/transformers/blob/main/examples/pytorch/question-answering/run_qa.py}. All experiments were run on 4 Nvidia GeForce RTX GPUs, and the training times varied from 10 GPU hours for the large models to 20 minutes for the tiny models.
The hyper parameter selection procedure was done according to the finetuning protocol described in \citet{Liu2019RoBERTaAR}. The hyper parameters\footnote{when using two learning rates the first learning rate used with seed 1-3 and the second 4-6.} and huggingface's datasets paths are described in Table \ref{tab:finetunings}.\begin{table*}
\centering
\resizebox{1.8\columnwidth}{!}{%
\begin{tabular}{lllrrrrr}
\toprule
\textbf{checkpoint} &\textbf{dataset} & \textbf{scheduler} & \textbf{warmup pr} & \textbf{batch} &  \textbf{lr} &  \textbf{epochs} & \textbf{seeds}\\
\midrule
prajjwal1/bert-tiny & glue/mnli & linear & 0.06 & 32 & 2e-5,3e-5 & 6 &  [1,...,6]\\
prajjwal1/bert-mini & glue/mnli & linear & 0.06 & 32 & 2e-5,3e-5 & 6 &  [1,...,6]\\
prajjwal1/bert-medium & glue/mnli & linear & 0.06 & 32 & 2e-5,3e-5 & 6 &  [1,...,6]\\
bert-base-uncased & glue/mnli & linear & 0.06 & 32 & 2e-5,3e-5 & 6 &  [1,...,6]\\
bert-large-uncased & glue/mnli & linear & 0.06 & 32 & 2e-5,3e-5 & 6 &  [1,...,6]\\
\midrule
roberta-base & glue/mnli & linear & 0.06 & 32 & 2e-5,3e-5 & 6 &  [1,...,6]\\
roberta-large & glue/mnli & linear & 0.06 & 32 & 2e-5,3e-5 & 6 &  [1,...,6]\\
\midrule
roberta-base & glue/qqp & linear & 0.06 & 32 & 2e-5,3e-5 & 6 &  [1,...,6]\\
roberta-large & glue/qqp & linear & 0.06 & 32 & 2e-5,3e-5 & 6 &  [1,...,6]\\
\midrule
electra-small & glue/qqp & linear & 0.0 & 32 & 2e-5 & 6 & [1,...,6]\\
electra-base & glue/qqp & linear & 0.0 & 32 & 2e-5 & 6 &  [1,...,6]\\
electra-large & glue/qqp & linear & 0.0 & 32 & 2e-5 & 6 &  [1,...,6]\\
\midrule
roberta-base & squad\_v2 & linear & 0.06 & 32 & 2e-5,3e-5 & 6 &  [1,...,6]\\
roberta-large & squad\_v2 & linear & 0.06 & 32 & 2e-5,3e-5 & 6 &  [1,...,6]\\
\midrule
electra-small & squad\_v2 & linear & 0.0 & 32 & 2e-5 & 6 & [1,...,6]\\
electra-base & squad\_v2 & linear & 0.0 & 32 & 2e-5 & 6 &  [1,...,6]\\
electra-large & squad\_v2 & linear & 0.0 & 32 & 2e-5 & 6 &  [1,...,6]\\
\bottomrule
\end{tabular}
}
\caption{All the hyper-parameters used in this work for fine-tuning the different models. Checkpoint is the hugginface hub identifier of the model.}
\label{tab:finetunings}
\end{table*}

The specifications of the different transformers model sizes mentioned in this paper, including their number of parameters are provided in Table \ref{tab:params_num}.
\begin{table}[t!]
\resizebox{1\columnwidth}{!}{%
\begin{tabular}{lccc}
\toprule
\textbf{Model} &\textbf{\# Layers} & \textbf{Hidden Dim} & \textbf{\# Parameters} \\
\midrule
Tiny  & 2 & 128 & 4M \\
Mini & 4 & 256 & 11M \\
Medium & 8 & 512 & 41.M \\
Small & 12 & 256 & 14M \\
Base & 12 & 768 & 110M \\
Large & 24 & 1024 & 355M \\
\bottomrule
\end{tabular}

}
\caption{Specifications of the different transformers
model sizes mentioned in this paper, including their
number of parameters.}
\label{tab:params_num}
\end{table}

\section{Datsets Details}
The standard datasets for each task, as well as HANS, were taken from the huggingface's dataset library. PAWS-QQP was generated by the code in \url{https://github.com/google-research-datasets/paws}.
\begin{table}[t!]
\resizebox{1\columnwidth}{!}{%
  \begin{tabular}{lrrr}
\toprule
\textbf{Dataset} & \textbf{Train-Size} &\textbf{Dev-Size} &\textbf{\ding{51}}\\
\midrule
MNLI & 392,702 & 9,815 &\\
HANS-Overlap \phantom{...}   &   10,000 & \phantom{......}10,000& 50\%\\
\midrule
QQP & 363,846 & 40,430& \\
PAWS-QQP & 11,988 & 677& 28.2\% \\
\midrule
SQuAD2.0 &130,319 & 11,873 &\\
ALSQA & - & 365& 50\%\\
\bottomrule
\end{tabular}
  }
  
  \caption{Details about the sizes of different datasets mentioned in this work. The percentage of positive labeled examples in each testing dev-set presented under the column titled with '\textbf{\ding{51}}'.}
  \label{tab:datasets_sizes}
\end{table}
\paragraph{Sizes of the Different Datasets} The details about the sizes of different datasets mentioned in this work can be found in Table \ref{tab:datasets_sizes}.

\section{On the Source of Generalization: Elaboration}
\label{sec:perplexity_app}

\begin{table}[t!]
\centering
\resizebox{1\columnwidth}{!}{%
\begin{tabular}{llrrrr}
\toprule
            &  &  \multicolumn{2}{c}{\textbf{Probable}} &  \multicolumn{2}{c}{\textbf{Improbable}} \\
       \textbf{Model} & \textbf{Size} &  \textbf{$PPL$} & $\Delta$ &  \textbf{$PPL$} & $\Delta$ \\
\midrule
    BERT &tiny &         13.9 &                &       75.6 &              \\
     &mini &          5.6 &         -59.7\% &       31.7 &        58.0\% \\
   &med&          3.2 &         -42.8\% &       16.0 &        49.5\% \\
    & base &          2.5 &         -21.8\% &       11.2 &        30.0\% \\
    &large &          2.4 &          -4.0\% &        9.5 &        15.1\% \\
   \midrule
  RoBERTa &base &          2.4 &                &       14.1 &              \\
 &large &          2.2 &          -8.3\% &        9.4 &        33.3\% \\
\midrule
       &rarity &         21.3 &                &      161.9 &              \\
\bottomrule
\end{tabular}
}
\caption{
The difference between the perplexity of pre-trained models on probable and improbable sentences. The relative improvement of every model size on top of its  between the perplexity of every model It can be seen that while the differences in perplexity between sizes of models are more then 3 times larger in rare sentences.
}
\label{tab:perplexity-full-results}
\end{table}

To approximate the distribution of the pre-training texts, we use a language models ensemble from different sizes. We use an ensemble to reduce the affect of the size of a single language model when computing sentence probabilities. In practice, we average the probabilities assigned by gpt2-large, gpt2-medium and gpt2-base. Using the probabilities assigned by the models ensemble 
we sort 3000 unseen sentences. 
The sampled sentences were not part of the models' training data, yet come from a similar distribution. In practice, we use the 1000 newest Wikipedia articles from the 15 June 2022 dump (the oldest article was created in the 13st of June 2022, well ahead the model's training time).
We then take the 20\% of sentences with the highest probability assigned by the language model ensemble, and mark them as \textit{probable} sentences and the lowest 20\% as \textit{improbable}. Then, we test the average perplexity that the different BERT and RoBERTa models assign to each group. Since these are MLM-based models, we compute the perplexity as defined by \cite{devlin-etal-2019-bert}, and later refered as pseudo perplexity \cite{salazar-etal-2020-masked} and calculated as exponent of the pseudo log likelihood \cite{wang-cho-2019-bert}. For a sentence $W=w_1,...,w_n$ and $P_{mlm}(W)_i$ (the probability for word $i$ in a sentence to be $w_i$ when unmasked by model $P_{mlm}$) the pseudo log likelihood is defined to be: $PLL(W) = \sum_{i=1}^nlog(P_{mlm}(W)_i)$. The corresponding perplexity is $PPL(W) = exp(PLL(W))$

The full results for the experiment are presented in Table \ref{tab:perplexity-full-results}.
\section{Additional Results}
In this section we describe additional results complementing the results  described in the paper. The results support the claims made in the main paper by supplying evident for the existence of the same trends in more model size variants and settings. 

\label{sec:add_res}

\subsection{Text Pair Classification}
\label{sec:add_res_pair}
Additionally to results presented in the main paper, we also report the full HEUR results on the five BERT models we consider (large, base, medium, mini and tiny), across epochs on HANS in Figure \ref{fig:hans_bert_std}. All models start with relativly high HEUR scores but during training HEUR scores gets better gradually. Additionally larger models tend to achive much better HEUR scores along training process.
more results are presented in Table \ref{tab:results_appendix}. Additionally the results of the models trained with span prediction objective show the same trend (Table \ref{tab:span-prediction-results})
Notice that in some experiments the HEUR does not go down, for example RoBERTa in the span setings and ELECTRA large in the binary settings. If we look closely on those cases we can see that there is earlier stage where the HEUR was lower but the development set performance was not much lower. In ELECTRA large in the middle of the first iteration the development set results are 1.1 points lower than the early stage achieving high result of 89.2 and from this point to the end of training the HEUR goes down by 3.7 points. this might indicate that the models may adopt the heuristic at very early stage than abounded it later on.

\subsection{Span Prediction Objective}
In this section we supply additional details about the answerability accuracy results of the models trained with the standard span prediction training on SQuAD2.0. The results on both the dev set and ALSQA test set are provided in Table \ref{tab:span-prediction-results}. We can see for example that ELECTRA-base has even higher dev set accuracy than ELECTRA-large, but the HEUR score of the larger model is much lower, indicating this model is less likely to use the lexical overlap heuristic. Comparison between the accuracy of ELECTRA early of different sizes on the consistent and inconsistent subsets of the data is visualized in Figure \ref{fig:alsqa_electra}.
\label{sec:add_res_span}

\begin{table}[t!]
\centering
\resizebox{1\columnwidth}{!}{%
\begin{tabular}{llrrrrr}
\toprule
\textbf{Model-Stage} &\textbf{Size} & \textbf{Dev}$\uparrow$ & \textbf{\ding{51}}$\uparrow$ & \textbf{\ding{55}}$\uparrow$ & \textbf{HEUR}$\downarrow$ & \textbf{$\Delta$}\\
\midrule
\multicolumn{6}{c}{\textbf{NLI}} \\
\midrule
 BERT early & tiny    & 64.0 & 99.9 & 0.2 &  99.7& \\
&   mini    & 71.8 & 95.1 & 3.8 &  91.3& \\
&med    & 79.0 & 84.6 & 13.4 &  71.2& \\
&base     & 83.8 & 97.3 & 3.6 &  93.7& \\
&large     & 84.9 & 91.1 & 27.6 & 63.5& \\
\midrule
BERT late & tiny  & 68.6 & 97.6 & 2.7 & 94.9 & -4.8\\
    &  mini& 76.0 & 63.7 & 35.3 & 28.4 & -42.8\\
    & med  & 81.0 & 70.8 & 33.7 & 37.1 & -54.2\\
    & base & 84.5 & 62.0 & 43.5 & 18.5 & -75.2\\
    &  large& 85.8 & 82.6 & 68.0 & 14.6 & -48.9\\
\midrule
RoBERTa early & base & 85.9 & 99.4	 & 9.7 &  89.7& \\
     & large & 87.8 & 99.8 &	81.4 &  18.4& \\
\midrule
RoBERTa late & base & 87.1 & 98.0 & 76.8 & 21.2 &  -68.5\\
    &  large & 89.3  & 99.6	& 92.4 & 4.2 & -14.2\\
\midrule
\multicolumn{6}{c}{\textbf{Paraphrase}} \\
\midrule
RoBERTa early& base & 89.4 & 93.7	 & 9.3 & 84.4& \\
    & large & 89.1 & 95.0 &	19.0 & 76.0 & \\
\midrule
RoBERTa late & base &91.6 & 90.1 & 21.9 & 68.2 & -16.2\\
        & large & 91.9 & 94.8	& 23.9 & 70.9 & -5.1\\
\midrule
ELECTRA early & small & 87.7 & 93.2	 & 6.2 & 87.0& \\
        & base & 90.0 & 89.3 &	17.5 & 71.8 & \\
         & large  & 90.8 & 98.4 &	23.9 & 74.5 & \\
\midrule
ELECTRA late & small  &90.1 & 91.6 & 10.6 & 81.0& -6.0\\
        & base & 91.8 & 89.5	& 35.1 & 54.4 & -17.4\\
        & large  & 92.5 & 93.7 &	42.0 & 51.7 &  -22.8\\
\midrule
\multicolumn{6}{c}{\textbf{Answerability}}\\
\midrule
ELECTRA early & small & 72.3 &89.5 & 31.4  & 58.1& \\
        & base & 83.5 & 90.8 & 53.7 & 37.1 & \\
         & large  & 90.3 & 89.5 &	71.4 & 18.1  & \\
\midrule
ELECTRA late & small & 72.9 & 89.7 & 32.6 & 57.1 & -1.0\\
        & base & 83.5 & 90.5	& 54.6 & 35.9 & -1.2\\
        & large  & 91.0 & 91.1 & 72.3 & 18.8 &  0.7\\
\bottomrule

\end{tabular}
}
\caption{
Results on all tasks considered in this study. We include both the dev-set results on the in-domain dataset (Dev), and the HEUR columns are reported on the high lexical overlap diagnostic sets.
\ding{51} refers to the subset that is consistent with the heuristic whereas \ding{55} refers to the subset that is inconsistent with it.
}
\label{tab:results_appendix}
\end{table}



\begin{table}[t!]
\centering
\resizebox{1\columnwidth}{!}{%
\begin{tabular}{llrrrrr}
\toprule
\textbf{Model-Stage} &\textbf{Size} & \textbf{Dev}$\uparrow$ &\textbf{\ding{51}}$\uparrow$  & \textbf{\ding{55}}$\uparrow$  & \textbf{HEUR}$\downarrow$ & \textbf{$\Delta$}\\
\midrule

ELECTRA early & small & 70.5 & 86.6 & 44.0 & 42.0 & \\
              & base  & 82.7 & 92.3 & 55.7 & 36.6 & \\
              & large & 89.3 & 91.0 & 70.6 & 20.4 & \\
\midrule
ELECTRA late  & small & 76.4 & 82.0 & 56.0 & 26.0 & -16.0\\
              & base  & 85.1 & 91.5 & 60.3 & 31.2 & -5.4 \\
              & large & 91.6 & 89.4 & 77.7 & 11.7 & -8.7 \\
\midrule
RoBERTa early& base  & 84.8  & 86.3 & 67.4  & 18.9 & \\
             & large & 88.8  & 86.6 & 75.4 & 11.2 & \\
\midrule
RoBERTa late & base  & 86.2  & 87.1 & 66.9  & 20.2 & 1.3\\
             & large & 90.2  & 88.7 & 74.9 & 13.8 & 2.6\\
\bottomrule

\end{tabular}
}
\caption{
Answerability results on ALSQA for models trained on SQuAD2.0 in span based finetuning. See Table \ref{tab:results} for the columns descriptions.}
\label{tab:span-prediction-results}
\end{table}

\end{document}